\documentclass[runningheads]{llncs}

\usepackage[mobile]{eccv}

\usepackage{eccvabbrv}

\usepackage{graphicx}
\usepackage{booktabs}

\usepackage[accsupp]{axessibility}  

\usepackage[pagebackref,breaklinks,colorlinks,citecolor=eccvblue]{hyperref}

\usepackage{orcidlink}

\begin{document}

\title{Isotropic3D: Image-to-3D Generation Based on a Single CLIP Embedding} 

\titlerunning{Isotropic3D}

\author{Pengkun Liu\inst{1,2,4}\and
Yikai Wang\inst{2}\textsuperscript{\dag} \and
Fuchun Sun\inst{2}\textsuperscript{\dag} \and
Jiafang Li\inst{4}\and
Hang Xiao\inst{1,2}\and
Hongxiang Xue\inst{1,2}\and
Xinzhou Wang\inst{3,2}
}

\authorrunning{Pengkun Liu et al.}

\institute{Academy for Engineering and Technology, Fudan University, Shanghai, China\and
Beijing National Research Center for Information Science and Technology (BNRist), State Key Lab on Intelligent Technology and Systems, Department of Computer Science and Technology, Tsinghua University, Beijing, China \and
Tongji University, Shanghai, China \quad $^4$ \ ShengShu, Beijing, China \\
}

\maketitle
{\let\thefootnote\relax\footnotetext{\noindent\textsuperscript{\dag} Corresponding Authors.}
\begin{center}
    \centering
    \captionsetup{type=figure}
    \includegraphics[width=1.\linewidth]{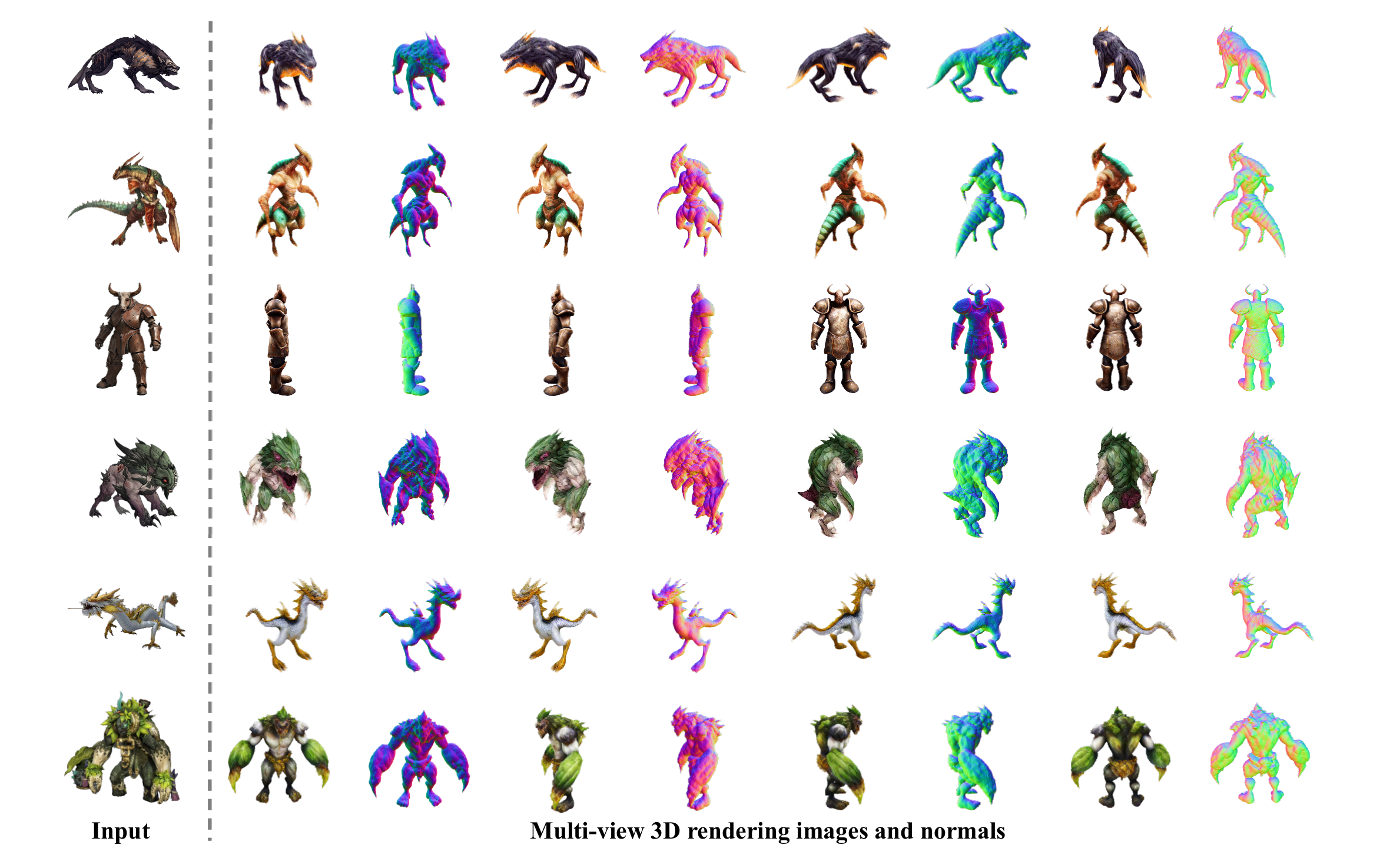}
    \captionof{figure}{Isotropic3D is a novel framework to generate multiview-consistent and high-quality 3D content from \textbf{a single CLIP embedding} of the reference image. Our method is proficient in generating multi-view images that maintain mutual consistency, as well as producing a 3D model characterized by symmetrical and neat content, regular geometry, rich colored texture, and less distortion, all while preserving similarity.}
    \label{fig:mvc123results}
\end{center}%

\begin{abstract}
Encouraged by the growing availability of pre-trained 2D diffusion models, image-to-3D generation by leveraging Score Distillation Sampling (SDS) is making remarkable progress. Most existing methods combine novel-view lifting from 2D diffusion models which usually take the reference image as a condition while applying hard L2 image supervision at the reference view. Yet heavily adhering to the image is prone to corrupting the inductive knowledge of the 2D diffusion model leading to flat or distorted 3D generation frequently. In this work, we reexamine image-to-3D in a novel perspective and present Isotropic3D, an image-to-3D generation pipeline that takes only an image CLIP embedding as input. Isotropic3D allows the optimization to be isotropic w.r.t. the azimuth angle by solely resting on the SDS loss. The core of our framework lies in a two-stage diffusion model fine-tuning. Firstly, we fine-tune a text-to-3D diffusion model by substituting its text encoder with an image encoder, by which the model preliminarily acquires image-to-image capabilities. Secondly, we perform fine-tuning using our Explicit Multi-view Attention (EMA) which combines noisy multi-view images with the noise-free reference image as an explicit condition. CLIP embedding is sent to the diffusion model throughout the whole process while reference images are discarded once after fine-tuning. As a result, with a single image CLIP embedding, Isotropic3D is capable of generating multi-view mutually consistent images and also a 3D model with more symmetrical and neat content, well-proportioned geometry, rich colored texture, and less distortion compared with existing image-to-3D methods while still preserving the similarity to the reference image to a large extent. The project page is available at \url{https://isotropic3d.github.io/}. The code and models are available at \url{https://github.com/pkunliu/Isotropic3D}. 

\end{abstract}

\section{Introduction}
\label{sec:intro}

Generating novel 3D contents that resemble a single reference image plays a crucial role in 3D computer vision, widely applicable to animation production, game development, and virtual reality \cite{goodfellow2014generative, luo2021diffusion, mildenhall2021nerf, wang2021neus,hsiang2022ar,eswaran2022challenges}. Thanks to the rapid growth of diffusion models in denoising high-quality images, there emerges a novel 3D generation pipeline that further synthesizes 3D objects by optimizing any 2D image views based on Score Distillation Sampling (SDS), as initially designed by DreamFusion \cite{poole2022dreamfusion} and widely adopted in many follow-up works \cite{lin2023magic3d, wang2023prolificdreamer, shi2023mvdream, liu2023zero, tang2023make, qian2023magic123, liu2023syncdreamer, chen2024gsgen3d, long2023wonder3d, wang2023imagedream, wang2023animatabledreamer}. 

Specifically for the image-to-3D task, it is natural to apply SDS optimization on novel azimuth angles with additional hard L2 supervision so that the rendered image at the reference view complies with the reference image. 
Furthermore, it should be noted that these methods \cite{liu2023zero, tang2023make, qian2023magic123, liu2023syncdreamer} mostly concatenate the reference image latent to the input noisy latent directly. In this way, they make the synthesis view resemble the input view as much as possible. 
However, empirical results indicate that such a kind of pipeline usually leads to three issues: 
i) 3D distortion or flattening. The conditional diffusion model will be limited in its generation capability. The way of forced supervision deviates from the original intention of generation, causing the model to compromise on conditional images and leading to flat or distorted 3D generation frequently.
ii) Multi-face problem. Due to the self-occlusion and invisible area, the network needs to rely on illusions to generate novel views. Generating other views that closely resemble the input view is a common challenge.
iii) Multi-view inconsistency. The generated 3D content cannot remain consistent across different viewpoints. These methods can only ensure that the reference image is as consistent as possible with the generated novel views, but tend to be weak at constraining the strong consistency between the multiple generated views.





\begin{table*}[t]
  \centering
  \tabcolsep=0.4em
  \caption{\textbf{Overview of related works in image-to-3D generation.} Distinguishing from previous works (especially SDS-based image-to-3D methods), our \textbf{Isotropic3D} only takes an image CLIP embedding as input and gets rid of the $L_2$ supervision loss. }
  \label{tab:compmain}
  \begin{tabular}{@{}llllll@{}}
    \toprule
    Method           & Prompt     & 3D model & Input style       & $L_2$ loss &  SDS  \\
    \midrule
    Realfusion \cite{melas2023realfusion} & Image   & NeRF  & CLIP + Image & $\checkmark$ & $\checkmark$           \\
    Zero123 \cite{liu2023zero}  & Image       & SJC  & CLIP + Image    & $\checkmark$   & $\checkmark$      \\
    MakeIt3D \cite{tang2023make} & Image + Text   & NeRF  & CLIP + Image    & $\checkmark$   & $\checkmark$      \\
    Magic123 \cite{qian2023magic123}   & Image + Text     & NeRF  & CLIP + Image   & $\checkmark$  & $\checkmark$       \\
    Syncdreamer \cite{liu2023syncdreamer}        & Image / Text     & NeRF / NeuS & CLIP + Image    & $\checkmark$      & $\times$  \\
    Wonder3D \cite{long2023wonder3d}   & Image     & NeuS  & CLIP + Image     & $\checkmark$   & $\times$  \\
    \textbf{Our Isotropic3D}       & Image & {NeRF}  & \textbf{CLIP} & \textbf{$\times$} & \textbf{$\checkmark$}   \\
    \bottomrule
  \end{tabular}
  \vspace{-1 em}
\end{table*}

To better address these issues, recent works \cite{gu2023nerfdiff, tang2023mvdiffusion, tseng2023consistent, zhou2023sparsefusion, shi2023mvdream, long2023wonder3d,deng2023nerdi,kim2023datid, wang2023imagedream} strive to generate multi-view images from a single image using 2D diffusion models. A text-to-3D generation method
MVDream \cite{shi2023mvdream} proposes a multi-view diffusion model that can generate consistent images. 
It turns out that the consistency between generated views and the quality of novel views largely determines the geometry and texture of the 3D content generated. 


In contrast to existing SDS-based image-to-3D generation methods, we introduce \textbf{Isotropic3D} in this work, an image-to-3D generation pipeline that takes only an image CLIP embedding as input. It allows the optimization to be isotropic w.r.t. the azimuth angle since the SDS loss is uniformly applied without being corrupted by the additional L2 supervision loss. We provide a systematic comparison with typical image-to-3D methods in Table~\ref{tab:compmain}, where ours is unique regarding both the input style and loss.
The key idea of Isotropic3D is to leverage the power of the 2D diffusion model itself without compromising on the input reference image by adding hard supervision during the 3D generation stage. Concretely, to preliminarily enable the diffusion to have the capability of image-conditioning,  we first fine-tune a text-to-3D diffusion model with a substituted image encoder. We then propose a technique dubbed Explicit Multi-view Attention (EMA) which further fine-tunes the diffusion model with the combination of noisy multi-view images and the \textbf{noise-free reference image} as an explicit condition. CLIP embedding is sent to the diffusion model throughout the whole process while reference images are discarded once after fine-tuning. 


Naively, an image CLIP embedding preserves semantic meanings but lacks geometry structures and textural details. However, thanks to our designed techniques in Isotropic3D, as shown in \cref{fig:3dgeneration}, We demonstrate that even with a simple CLIP, our framework can still generate high-quality 3D models with rich color and well-proportioned geometry. We observe that our method is robust to the object pose of the reference image. Besides, there is still a large degree of consistency retained with the reference image.

To summarize the contribution of our paper as follows:
\begin{itemize}
    \item We propose a novel image-to-3D pipeline called Isotropic3D that takes only an image CLIP embedding as input. Isotropic3D aims to give full play to 2D diffusion model priors without requiring the target view to be utterly consistent with the input view.
    \item We introduce a view-conditioned multi-view diffusion model that integrates Explicit Multi-view Attention (EMA), aimed at enhancing view generation through fine-tuning. EMA combines noisy multi-view images with the noise-free reference image as an explicit condition. Such a design allows the reference image to be discarded from the whole network during the SDS-based 3D generation process.
    \item Experiments demonstrate that with a single CLIP embedding, Isotropic3D can generate promising 3D assets while still showing similarity to the reference image.
\end{itemize}


\section{Related Work}
\label{sec:relatedwork}

Our work focuses on 3D generation from a single image. In this section, we review the literature on 3D generative models and optimize-based 3D generation, which has achieved remarkable performance by utilizing the capability of diffusion architecture and pre-trained models.

\subsection{3D Generative Models}
Generative models such as variational autoencoders (VAEs) \cite{kingma2013auto}, generative adversarial networks (GANs) \cite{goodfellow2014generative}, and diffusion models (DMs) \cite{ho2020denoising} have achieved remarkable success in the field of 2D generation. Recently, research \cite{henderson2020learning,henderson2020leveraging,gao2022get3d,muller2023diffrf,anciukevivcius2023renderdiffusion,baillif2023deep,dundar2023fine} has extended its application to 3D generation. AutoSDF \cite{mittal2022autosdf} applied VQ-VAE \cite{van2017neural} to project high-dimensional continuous 3D shapes into low-dimensional latent space and combined it with a transformer to complete the conditional generation task. By integrating 3D scenes into GANs, the new model \cite{niemeyer2021giraffe,xue2022giraffe,wu2016learning,tan2022volux,xu2023discoscene,xie2023high} exhibits improved capability in generating images of higher quality and controllability.

Building upon the 2D diffusion model, 3D-aware methods like \cite{xiang20233d} have reformulated the task of 3D perceptual image generation. They approach it by generating a multi-view 2D image set, followed by developing a sequential unconditional-conditional process for multi-view image generation.
DreamFields \cite{jain2022zero} combined neural rendering with image and text representations to synthesize diverse 3D objects from natural language prompts independently. The model can generate the geometry and color of a variety of objects without 3D supervision. Based on the DreamFields \cite{jain2022zero}, DreamFusion \cite{poole2022dreamfusion} used the Imagen text-to-image diffusion model \cite{saharia2022photorealistic} to replace the CLIP model \cite{radford2021learning}, which enhanced the quality of 3D content derived from natural language and demonstrated the feasibility of generating a 3D model from a single 2D image.

\subsection{Optimize-based 3D Generation}
Dreamfusion \cite{poole2022dreamfusion} proposed Score Distillation Sampling (SDS) to address 3D data limitations, which has driven the recent development of 2D lifting methods \cite{xu2023neurallift,tang2023make,shen2023anything,ruiz2023dreambooth,raj2023dreambooth3d,liu2023zero}. 

Zero123 \cite{liu2023zero} proposed a single-view 3D generation framework, that leveraged geometric prior knowledge learned from natural images using large-scale diffusion models to generate novel views. The generative model, when coupled with NeRF \cite{wang2021nerf}, is capable of effectively modeling 3D scenes from single-view images.
MakeIt3D \cite{tang2023make} designed a universal 3D generation framework that utilized diffusion priors as 3D perception supervision in a coarse-to-fine manner to create high-fidelity 3D content from a single image. 
Although achieving high-quality and high-fidelity target generation without suffering from the limitations of 3D data, these models occurred inconsistent multi-view generation. To cope with the problem, some methods \cite{lin2023consistent123,weng2023consistent123,long2023wonder3d,liu2023syncdreamer,ye2023consistent} try to add conditional constraints to supervise the image consistency in the process of applying the 2D diffusion model to generate multi-view images. 
Wonder3D \cite{long2023wonder3d} enhanced information exchange among different views through the introduction of cross-domain attention, which is proficient in generating multi-view images that preserve both semantic and geometric coherence.
MVDream \cite{shi2023mvdream} integrates 2D image generation with 3D data consistency, guiding 3D generation through a multi-view prior. This approach not only preserves the generalization capability of 2D generation but also enhances the performance of 3D tasks. 
As a concurrent effort, Imagedream \cite{wang2023imagedream} necessitates highly-matched image-text correspondence. Nevertheless, well-designed text prompts also struggle to accurately describe image information. It also introduces a new set of MLPs inserted in the MVDiffusion side, which increases the difficulty of model training. In contrast, Isotropic3D only requires a single image as input to the model, eliminating the need for text prompts. Additionally, we employ the pre-trained CLIP model directly as the image encoder and keep it frozen throughout the training process.


\section{Method}
\label{sec:method}
We propose Isotropic3D, as shown in \cref{fig:pipline}, which is an image-to-3D generation pipeline that takes only an image CLIP embedding as input and allows the optimization to be isotropic w.r.t. the azimuth angle by solely resting on the SDS loss.
Isotropic3D is composed of two parts: 
i) \textbf{View-conditioned muti-view diffusion model}. A framework with Explicit Multi-view Attention (EMA) is used to generate diverse but high-quality consistent multi-view images.
ii) \textbf{Neural Radiance Field (NeRF)}. A 3D network yields high-quality 3D content optimized by rendered images via Score Distillation Sampling (SDS).


\begin{figure}[tb]
  \centering
  \vspace{-0.4cm}
  \includegraphics[width=1.\linewidth]{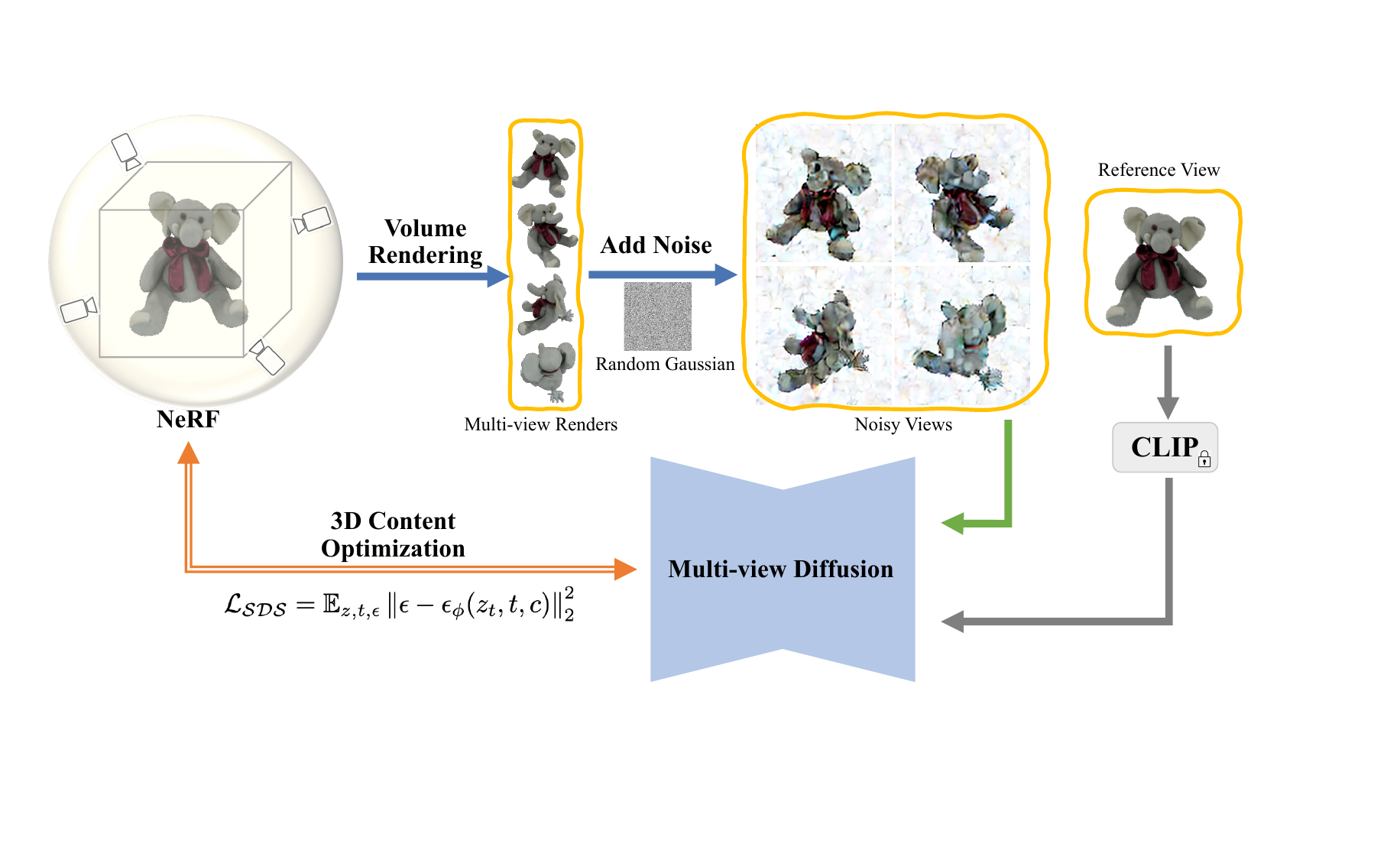}
  \vspace{-2.0cm}
  \caption{The pipeline of Isotropic3D. Neural Radiance Field (NeRF) utilizes volume rendering to extract four orthogonal views, which are subsequently augmented with random Gaussian noise. These views, along with noise-free reference images, are then transferred to a multi-view diffusion model for predicting added noise. Note that, we \textbf{set the timestep $t$ to zero} at the corresponding position of noise-free reference images. The framework that generates consistent multi-view images from only a single CLIP embedding can be aligned with the input view while retaining the consistency of the output target view. Finally, NeRF yields high-quality 3D content optimized by rendered images via Score Distillation Sampling (SDS). $\mathcal{L_{SDS}}$ can refer to \cref{eq:sds}. }
  
  \label{fig:pipline}
\end{figure}

\subsection{Motivation}
\label{ssec:motivation}
In order to align the reference image and target images, Zero123 adopts two strategies: one concatenates the latent target view encoded by VAE \cite{kingma2013auto} with the input view latent on the channel, and the other takes the CLIP embedding of the reference image as conditional information. 

Some recent works improve on this basis, consistent123 \cite{weng2023consistent123} and Zero123plus \cite{shi2023zero123++} apply to share self-attention mechanism which appends a self-attention key and value matrix from a noisy input view image to the corresponding attention layer. The same level of Gaussian noise as the target view is added to the input view image and then denoising via the UNet network together with the noisy target view. 
However, we found that existing methods combine novel-view lifting from 2D diffusion models which usually take the reference image as a condition while applying hard L2 image supervision at the reference view. 

Unlike the previous 3D generation with complex strong constraints, our goal is to generate more regular geometry, naturally colored textures, and less distortion with only an image CLIP embedding as input. At the same time, 3D content still preserves the similarity to the reference image to a large extent.
Therefore, we present Isotropic3D, an image-to-3D generation pipeline that takes only an image CLIP embedding as input. Isotropic3D allows the optimization to be isotropic w.r.t. the azimuth angle by solely resting on the SDS loss. 

\subsection{View-Conditioned Multi-view Diffusion}
\label{ssec:mvd}
\begin{figure}[tb]
  \centering
    \includegraphics[width=1.0\linewidth]{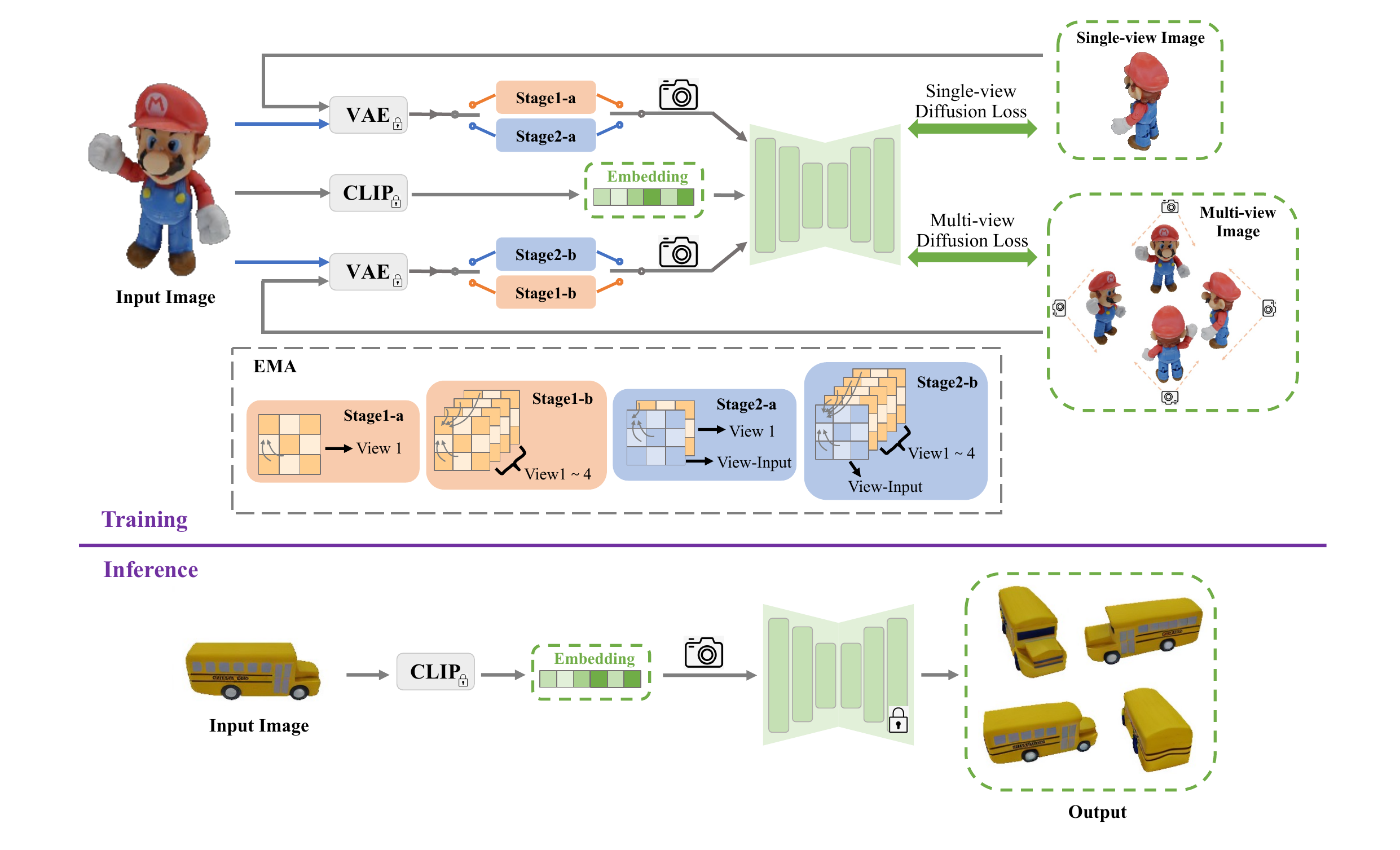}
    \vspace{-0.6cm}
  \caption{View-Conditioned Multi-view Diffusion pipeline. Our training process is divided into two stages. In the first stage ({Stage1}), we fine-tune a text-to-3D diffusion model by substituting its text encoder with an image encoder, by which the model preliminarily acquires image-to-image capabilities. {Stage1-a} and {Stage1-b} are the single-view diffusion branch and the multi-view diffusion branch for the first stage respectively. In the second stage ({Stage2}), we perform fine-tuning multi-view diffusion model integrated Explicit Multi-view Attention (EMA). EMA combines noisy multi-view images with the noise-free reference image as an explicit condition. {Stage2-a} and {Stage2-b} are diffusion branches for the second stage.
   During inference, we {only} need to send the {\bf CLIP embedding} of the reference image and camera pose to generate consistent high-quality images from multiple perspectives.}
  \label{fig:mvdpipline}
\end{figure}
\textbf{Architecture.} 
Given a reference image $y \in \mathbb{R}^{1 \times H \times W \times C} $ as model input view, our method is to generate multi-view images $x \in \mathbb{R}^{N \times H \times W \times C}$ from $N$ different viewpoints aligned with input view and keep consistent to each other. The VAE encoder is denoted as $\mathcal{E}$. The latent vector of reference image can be written as $z^v=\mathcal{E}(y)$. The camera parameters of different viewpoints is $\pi = \{\pi_1, \pi_2, ..., \pi_N \}$.
We denote joint probability distribution as $p(x,y) = p_\theta (x | y) p_\theta(y)$. 
In multi-view diffusion, this distribution can be written as 
\begin{equation}
    p(x^{(1:N)},y) := p_\theta(x^{(1:N)} | y).
\end{equation}
Therefore, the reverse process of the view-conditioned multi-view diffusion model can be extended. We can formulate this process as

\begin{equation}
    p_\theta(\boldsymbol{x}^{1:N}_{0: T}, c) = p(x_T^{1:N}, c) \prod_{t=1}^T p_\theta(\boldsymbol{x}^{1:N}_{t-1} \mid \boldsymbol{x}^{1:N}_t, c),
\label{eq:multiviewreverse}
\end{equation}
where $p(x_T^{1:N}, c)$ represents Gaussian noises, while $p_\theta(\boldsymbol{x}^{1:N}_{t-1} \mid \boldsymbol{x}^{1:N}_t, c)$ denotes a Gaussian distribution. Here $t$ is the time step, and $c$ encompasses condition information, comprising the reference image $y$ and camera parameters $\pi$.

To inherit the performance of MVDream \cite{shi2023mvdream}, the view-conditioned muti-view diffusion model is designed with two branches: single-view generation and multi-view generation, as shown in \cref{fig:mvdpipline}. 
Single-view generation branch receives input from a pair of random perspectives. The purpose of this is to preserve the model's ability to generate arbitrary views. 
Multi-view generation branch takes one of the random perspectives as input view, but the outputs are from four perspectives. Through such supervised training, we lay the foundation for ensuring that the model can generate arbitrary perspectives while ensuring consistency between generated views. We will introduce the data preparation in \cref{ssec:implementation}.

\textbf{Explicit multi-view attention (EMA).}
\label{ssec:refattn}
\begin{figure}[tb]
  \centering
    \includegraphics[width=0.7\linewidth]{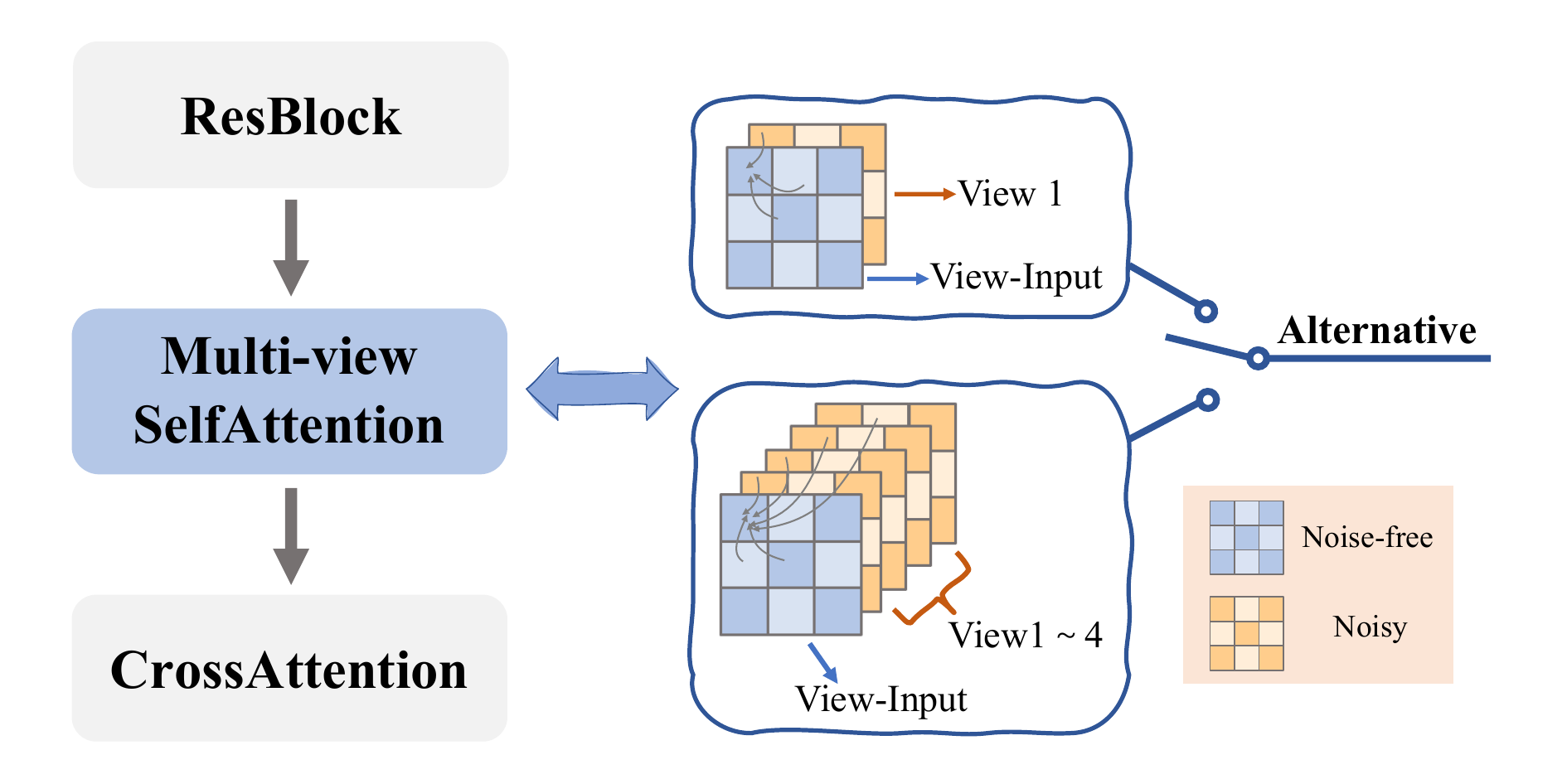}
    \vspace{-0.2cm}
   \caption{Illustration of the Explicit Multi-view Attention (EMA). ``{View-Input}'' is a feature map of the noise-free reference image. ``{View 1}'' and ``{View 1 $\sim$ 4}'' are feature maps of noisy rendered views. ``{Alternative}'' means a 30\% chance of using single-view diffusion (Stage2-a) and a 70\% chance of training with the multi-view diffusion branch (Stage2-b). 
   }
   \vspace{-0.1 in}
   \label{fig:MVD}
\end{figure}
Achieving high-quality and consistent target views is fundamental to generating regular geometry and detailed texture. To this end, we design a new attention mechanism called Explicit Multi-view Attention (EMA), as shown in \cref{fig:MVD}.

In contrast to Zero123 \cite{liu2023zero}, MVDream \cite{shi2023mvdream} and Wonder3D \cite{long2023wonder3d}, our Explicit Multi-view Attention concatenates the noise-free reference image feature with the noisy image latent/latents as the network input. At the same time, the corresponding timesteps $t^v$ and Gaussian noise $\epsilon^v$ of the noise-free reference image are set to \textbf{0}. The noisy latent vector $z_t$ can be written as
\begin{equation}
\boldsymbol{z}_t=\sqrt{\bar{\alpha}_t} \boldsymbol{z}+\sqrt{1-\bar{\alpha}_t} \boldsymbol{\epsilon},
\end{equation}
and thus the noise-free latent vector $z^{v}_{t}$ is denoted as
\begin{equation}
\begin{aligned}
\boldsymbol{z}^v_t &= 
\boldsymbol{z}^v
    & s.t. \sqrt{\bar{\alpha}_t} = 1, t = t^v=0, \epsilon = \epsilon^v = \boldsymbol{0},
\end{aligned}
\end{equation}
where $\bar{\alpha}_t$ is variance schedule \cite{ho2020denoising}, $\boldsymbol{\epsilon} \sim \mathcal{N}(\boldsymbol{0}, \boldsymbol{I})$. The purpose is that our target view can clearly capture the characteristic details of the input view during the self-attention process of the model. 

\textbf{Optimazation.}
The core of our Isotropic3D lies in this two-stage view-conditioned multi-view diffusion model fine-tuning. The first stage aims to transform the model from text-to-image to image-to-image. We fine-tune a text-to-3D diffusion model by substituting its text encoder with an image encoder, by which the model preliminarily acquires image-to-image capabilities. Following the above discussion, the optimization objective for the first stage can be denoted as
\begin{equation}
    \mathcal{L_{MV}} = \mathbb{E}_{z,t,\pi,\epsilon} \left \|\epsilon_\theta(z_t, t, \pi) - \epsilon \right\|_2^2,
\end{equation}
where $\epsilon_\theta$ signifies the multi-view diffusion process targeted at denoising the noisy latent variable $z_t$. The variable $t$ indicates the timestep, and the parameter $\pi$ pertains to the camera parameters.

In the second stage, we perform fine-tuning using Explicit Multi-view Attention (EMA), which integrates noisy multi-view images with the noise-free reference image as an explicit condition. To prevent the model from interfering with the consistent relationship of target views, we opt for the prediction noise associated with the target views rather than the prediction noise linked to the reference image. It allows the model only to learn the consistency of the target views and ignores the input view. This strategy enables the model to focus solely on learning the consistency of the target view while disregarding the reference view. The optimization objective for this process can be expressed as
\begin{equation}
    \mathcal{L_{E-MV}} = \mathbb{E}_{z^v,z,t_v,t,\pi_v,\pi,\epsilon} \left \|\epsilon_\theta((z^v_t \oplus z_t), (t_v \oplus t), (\pi_v \oplus \pi)) - \epsilon \right\|_2^2,
\label{eq:multiviewreverse}
\end{equation}
where noise-free latent $z^v$ is derived from the reference image, which is encoded by a Variational Autoencoder (VAE). The variable $t_v$ indicates the timestep set to 0. The parameter $\pi_v$ specifies the camera parameters when both elevation and azimuth are set to 0.
We performed explicit multi-view attention on both single-view generation and multi-view generation branches.

\subsection{NeRF Optimization Stage} 

Given a Nerual Radiance Fields $\mathcal{G}$, we can randomly sample a camera pose parameter and render a corresponding view $x$. The rendered view can be denoted as $x=\mathcal{G}(\theta)$.
Dreamfusion \cite{poole2022dreamfusion} proposes to use a 2D diffusion model prior to optimizing the NeRF via score distillation sampling (SDS) loss.
With the help of an image-to-image 2D diffusion model, a target view is generated when the loss function is minimized, and then the parameter $\theta$ is optimized so that $x$ looks like a sample of the frozen diffusion model. The SDS loss is formulated as

\begin{equation}
    \mathcal{L_{SDS}}= \mathbb{E}_{z,t,c,\epsilon} \left \|\epsilon - \epsilon_\phi(z_t,t,c)\right\|_2^2,
    \label{eq:sds}
\end{equation}
where $z$ is the latent rendered by NeRF with added noise, $\epsilon$ refer as the Gaussian noise, $c$ is composed of camera parameters $\pi$ and the reference image $y$.
For NeRF optimization, we solely utilize SDS and orientation loss \cite{poole2022dreamfusion} which encourage normal vectors of the density field facing toward the camera when they are visible. 
The orientation loss \cite{poole2022dreamfusion} is written as
\begin{equation}
\mathcal{L}_{\text {orient }}=\sum_i \operatorname{stop\_grad}\left(w_i\right) \max \left(0, \boldsymbol{n}_i \cdot \boldsymbol{v}\right)^2,
\end{equation}
where $w_i$ is rendering weights, and the direction of the ray is denoted as $\boldsymbol{v}$. For regularizing geometry, we choose point lighting and soft shading. We empirically set the guidance scale to 10 which is the same as during multi-view diffusion training. We define our total loss function as
\begin{equation}
\mathcal{L}=\lambda_e \mathcal{L_{SDS}} + \lambda_o \mathcal{L}_{\text {orient }},
\end{equation}
where $\lambda_e$ and $\lambda_o$ are loss weights.

\section{Experiments}
\label{sec:experiments}
\begin{figure}[tb]
  \centering
    \includegraphics[width=1.\linewidth]{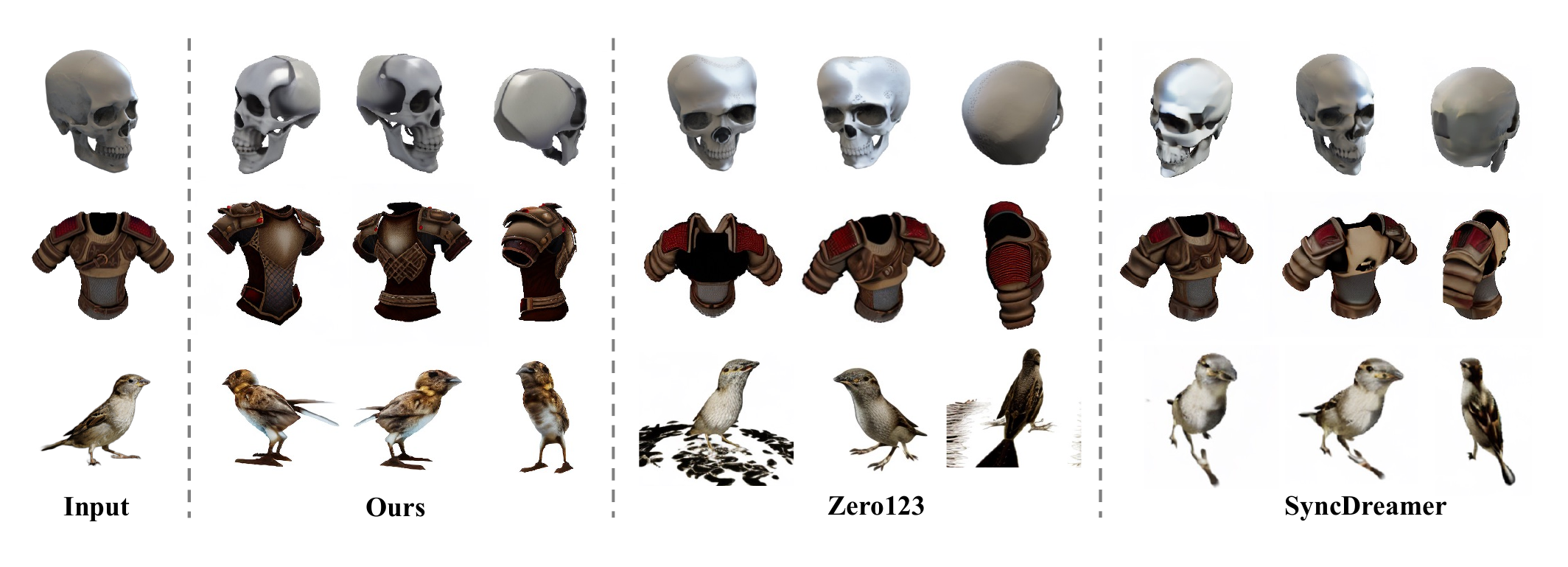}
  \caption{Qualitative comparison of synthesizing novel views with baseline models \cite{liu2023zero, liu2023syncdreamer} on GSO \cite{downs2022google} and randomly collected images.}
  \vspace{-0.1 in}
  \label{fig:novelview}
\end{figure}
We provide implementation details in \cref{ssec:implementation} and evaluate novel view synthesis with baselines in \cref{ssec:novelview}. Furthermore, we compare the ability of 3D generation with image-to-3D methods based on SDS in \cref{ssec:3dgen}. To assess EMA module and the advantages of Isotropic3D with a single embedding as input, we conduct an ablation study in \cref{ssec:ablation}.

\subsection{Implementation Details}
\label{ssec:implementation}
\textbf{Datasets preparation.} The Objaverse dataset \cite{deitke2023objaverse} is a large-scale dataset comprising over 800k annotated 3D objects. We fine-tune our model using this extensive 3D dataset. Following the rendering settings of Syncdreamer \cite{liu2023syncdreamer}, all images are resized to $256 \times 256$, with the background reset to white. The camera distance is set to 1.5, and the lighting is randomized HDRI sourced from Blender. We render both a random view set and a fixed perspective set. Each object is rendered with 16 views for both the random and fixed view sets. In the random view set, the elevation range of images is [$-10^\circ$, $40^\circ$], while the azimuths remain constant. For the fixed view set, the azimuths of target views are evenly spaced within the range [$0^\circ$, $360^\circ$], with a fixed elevation of $30^\circ$. Additionally, we utilize the Google Scanned Objects (GSO) dataset \cite{downs2022google} and randomly collected images to evaluate the performance of our method.

\textbf{Training procedure.} The multi-view generation framework comprises two main branches: single-view diffusion and multi-view diffusion. During tuning, We have a 30\% chance of using single-view diffusion and a 70\% chance of training with the multi-view diffusion branch. 
The whole tuning process is divided into two stages. In the first stage, we train an image-to-image model from the text-to-image model called MVDream\cite{shi2023mvdream} and keep the same settings of optimizer and $\epsilon$-prediction. The training with a batch size of 768 takes about 7 days.
In the second stage, we incorporate the explicit attention mechanism to multi-view diffusion model and fine-tune full UNet. The batch size is set to 128 and the training time takes about 1 day. All training is done on 8 Nvidia A800 GPUs. 
After tuning, Isotropic3D demonstrates the capability to generate multi-view images with only a single CLIP embedding that exhibit mutual consistency and a 3D model characterized by more well-proportioned geometry and colored texture. The 3D generation typically takes around 1 hour on a single GPU.

\begin{figure}[tb]
  \centering
    \includegraphics[width=1.\linewidth]{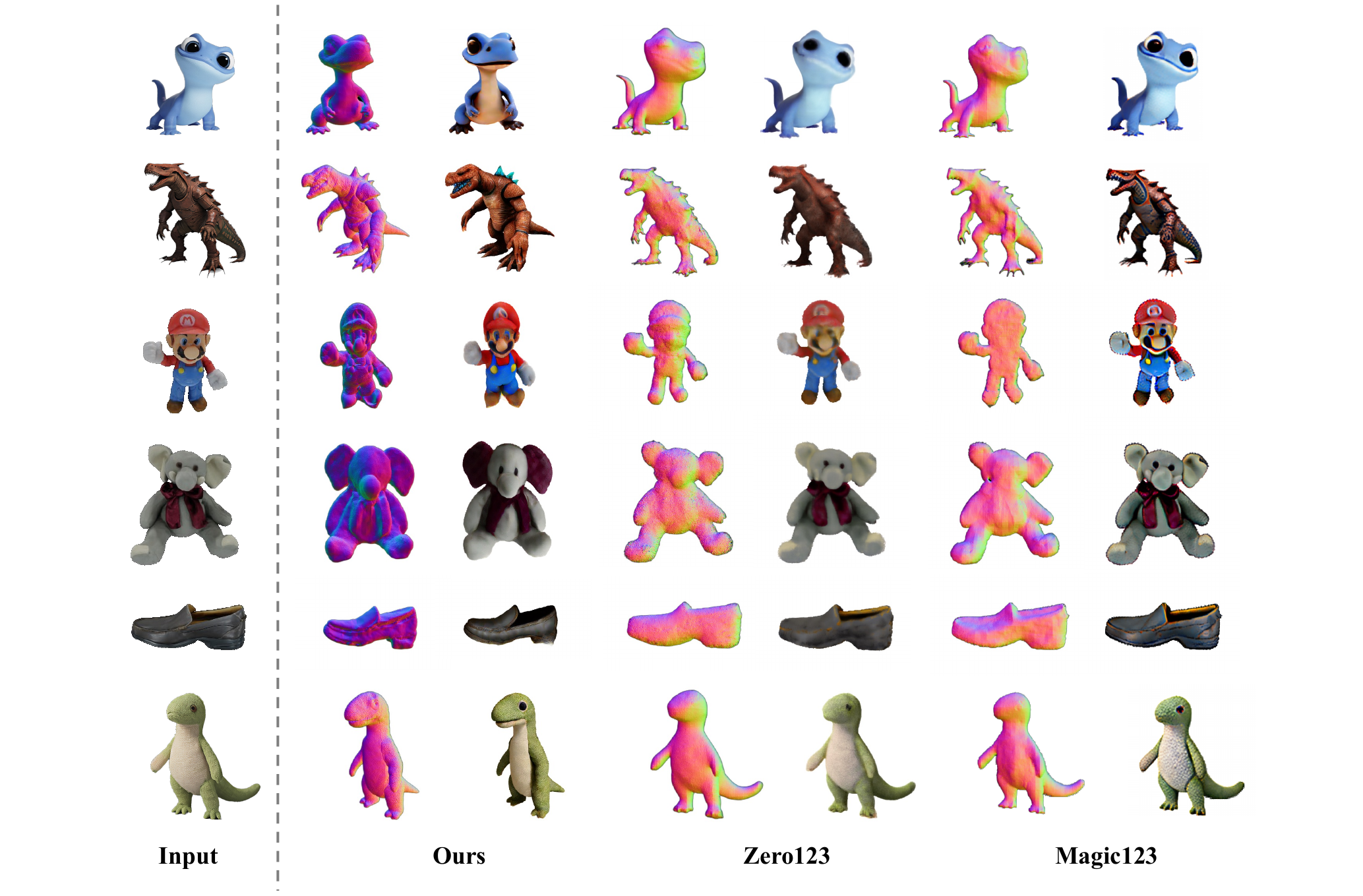}
  \caption{Qualitative comparisons of 3D Generation with baseline models. We conducted verification on GSO \cite{downs2022google} and randomly collected images. 
  {\bf Isotropic3D} is capable of generating more regular geometry, detailed texture, and less flat compared with Zero123 and Magic123. A video of this result is available at \url{https://isotropic3d.github.io/}.}
  \vspace{-0.1in}
  \label{fig:3dgeneration}
\end{figure}

\textbf{Baselines.} We reproduce and compare the diffusion-based baseline methods including Zero123 \cite{liu2023zero}, MakeIt3D \cite{tang2023make}, Magic123 \cite{qian2023magic123}, Syncdreamer \cite{liu2023syncdreamer}. Zero123\cite{liu2023zero} can generate novel-view images of an object from a single-view image. In addition, the model can also be combined with NeRF to perform 3D reconstruction of objects. MakeIt3D \cite{tang2023make} leverage prior knowledge from a well-trained 2D diffusion model to act as 3D-aware supervision for high-quality 3D creation. Magic123 \cite{qian2023magic123} adopts a two-stage optimization framework to generate high-quality 3D content by combining 2D prior and 3D prior. Although Zero123 \cite{liu2023zero} can generate high-quality novel images, there are still difficulties in maintaining consistency in multi-view images.
Therefore, SyncDreamer \cite{liu2023syncdreamer} is proposed that generates consistent images from multiple views by utilizing a 3D-aware feature attention mechanism.


\subsection{Novel View Synthesis}
\label{ssec:novelview}
Two factors affect the quality of 3D content generation: one is view consistency, and the other is the quality of new view generation. We compare the synthesis quality of novel views with the baseline models. The qualitative results are shown in \cref{fig:novelview}. We can find that the images generated by zero123 \cite{liu2023zero} maintain consistency with the reference images, but there is a lack of consistency between the generated views. Syncdreamer \cite{liu2023syncdreamer} designed the volume attention module to enhance the consistency between views, but its generated results appeared to be pathological views when far away from the reference image and were inconsistent with other generated views.
Compared with above methods, our model can ensure high-quality novel views and is aligned with the semantics of input views.

\subsection{3D Generation}
\label{ssec:3dgen}
\begin{figure}[t]
  \centering
  \includegraphics[width=0.9\linewidth]{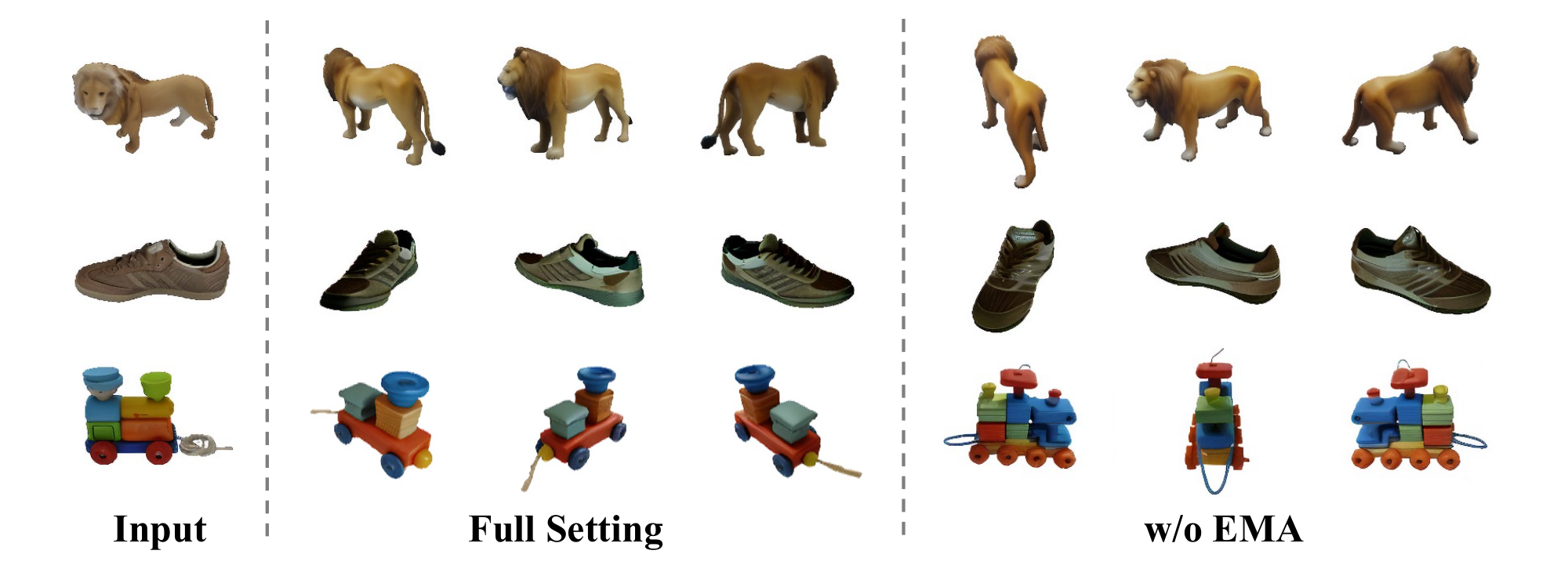}
  \caption{Ablation studies on Explicit Multi-view Attention. }
  \label{fig:ablation}
  \vspace{-0.1in}
\end{figure}
We evaluate the geometry quality generated by different methods. The qualitative comparison results are shown in \cref{fig:3dgeneration}. 
For each instance, we only optimize NeRF once via SDS loss, and the 3D contents shown in \cref{fig:3dgeneration} are NeRF renderings.
For a fair comparison, we perform the first stage of Zero123 and Magic123.
For Zero123 \cite{liu2023zero} and Magic123 \cite{qian2023magic123}, their normal is rougher and the geometry is smoother. 
In contrast, our method performs well in generating 3D models. We do not require the generated content to be aligned completely with the input view, only semantic consistency with the reference image. We can find that our 3D assets maintain high-quality and detailed geometry in texture. Isotropic3D is capable of generating regular geometry, colored texture, and less distortion from a single CLIP embedding compared with existing image-to-3D methods.

\subsection{Ablation Study}
\label{ssec:ablation}
\textbf{Explicit multi-view attention.} To verify the effectiveness of our Explicit Multi-view Attention (EMA), we compared the method using multi-view attention proposed by MVDream \cite{shi2023mvdream}, which is also used in Wonder3D \cite{long2023wonder3d}. The qualitative results are shown in \cref{fig:ablation}.  
We can find that after the second stage of fine-tuning, the lion's leg posture in the first row is more similar to the reference image. At the same time, the texture details of the shoes in the second row are more similar to the reference image. Using explicit multi-view attention can improve the similarity between the target views and the input view without changing the consistency of the target views.

\begin{figure}[t]
  \centering
  \includegraphics[width=1.\linewidth]{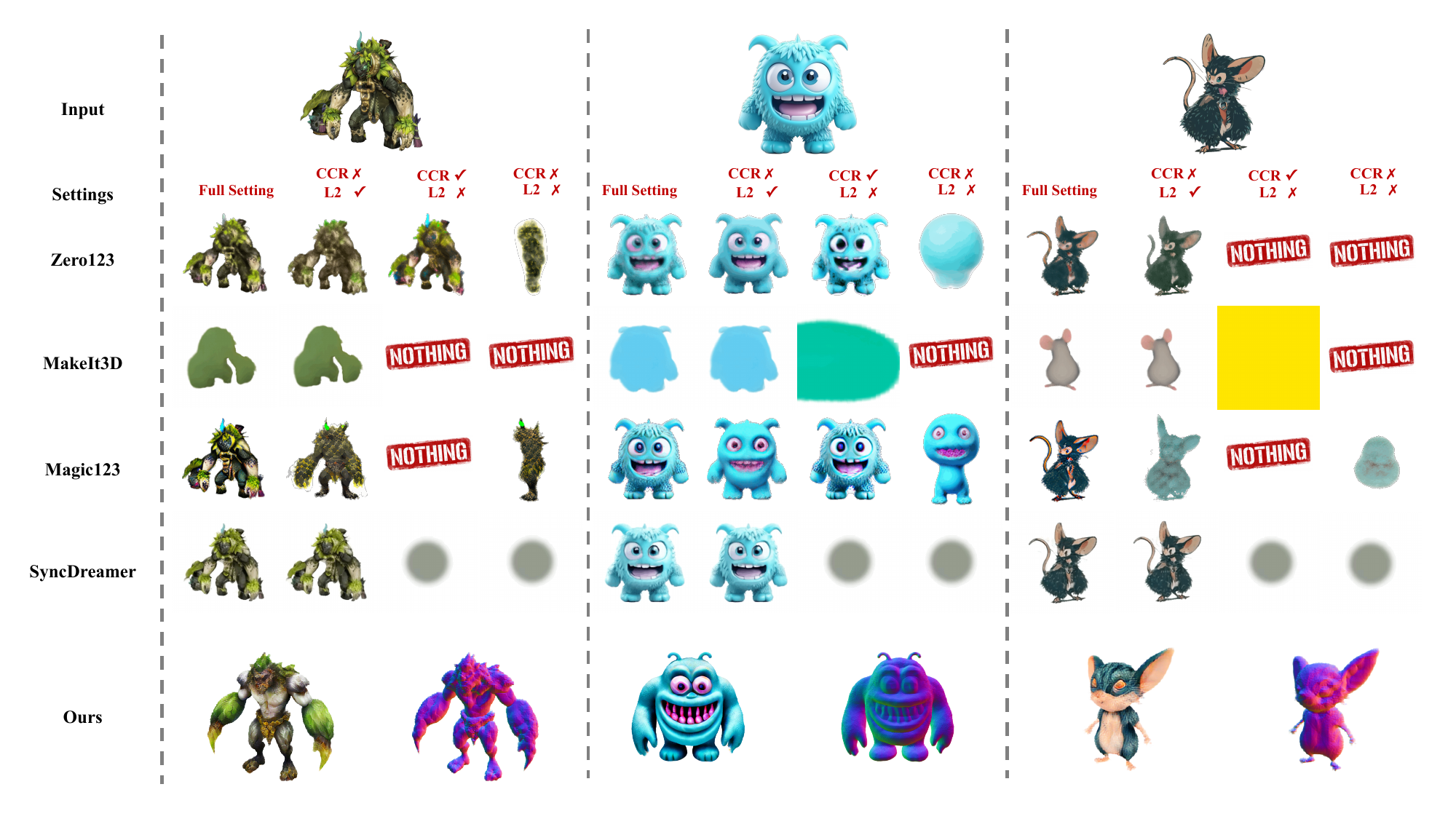}
    \vspace{-0.3 in}
  \caption{Qualitative comparisons in different settings. \emph{CCR} is denote as {\bf c}hannel-{\bf c}oncatenate {\bf r}eference image. \emph{NOTHING} means that it does not generate anything. A video of this result is available at \url{https://isotropic3d.github.io/}.}
  \vspace{-0.1 in}
  \label{fig:morecomparison}
\end{figure}

\textbf{Comparsion results with other methods on different settings.} As shown in \cref{fig:morecomparison}, we compare Isotropic3D with Zero123 \cite{liu2023zero}, MakeIt3D \cite{tang2023make}, Magic123 \cite{qian2023magic123} and Syncdreamer \cite{liu2023syncdreamer} under different settings:
\begin{itemize}
    \item {\bf Using full setting.} All settings are set according to the original parameters of the model. Here we use threestudio \footnote{https://github.com/threestudio-project/threestudio} library for Zero123 and Magic123. MakeIt3D and Syncdreamer use official implementation.
    
    \item {\bf Removing channel-concatenate reference image.} Zero123, Magic123 and Syncdreamer concatenate the reference image with the noisy image in the channel dimension. MakeIt3D does not use channel-concatenate reference image. In order to ensure that the input dimension of the network remains unchanged, we replace the position corresponding to the reference image with a zero-like matrix of equal dimensions.

    \item {\bf Removing $L_2$ loss supervision.} Zero123, MakeIt3D, Magic123 and Syncdreamer use reference image for L2 supervision. We reset all loss weights related to the reference image to zero, including RGB loss, depth loss, and mask loss.

    \item {\bf Removing channel-concatenate reference image and $L_2$ loss supervision together.} Removing channel-concatenate reference image and $L_2$ loss supervision together means {\bf generating 3D content with a single CLIP embedding}. Note that MakeIt3D does not use channel-concatenate reference image, we only remove $L_2$ loss supervision.
    
\end{itemize}

In \cref{fig:morecomparison}, existing image-to-3D methods rely so much on the reference image that they are almost impossible to generate a complete 3D object. When we remove the channel-concatenate reference image, the texture of the 3D model generated by Zero123, Magic123 and Syncdreamer will be reduced. MakeIt3D does not generate properly in most cases. After removing  $L_2$ loss supervision, MakeIt3D and Syncdreamer can not generate anything at all. When removing channel-concatenate reference image and $L_2$ loss supervision together, it means that only using {\bf a single CLIP embedding} to generate 3D models.
Only Zero123 and Magic123 can generate low-quality objects without regular geometry and clear texture. MakeIt3D and Syncdreamer can not generate anything completely in our test cases.
In comparison, our method can generate multi-view mutually consistent images and high-quality 3D models with only an image CLIP embedding as input.

\section{Conclusion}
\label{sec:conclusion}
In this paper, we propose Isotropic3D, a new image-to-3D pipeline to generate high-quality geometry and texture only from an image CLIP embedding. Isotropic3D allows the optimization to be isotropic w.r.t. the azimuth angle by solely resting on the SDS loss.
To achieve this feat, we fine-tune a multi-view diffusion model in two stages, which aims to utilize the semantic information of the reference image but does not require it to be completely consistent with the reference image, thereby preventing the diffusion model from compromising the reference view. 
Firstly, we perform fine-tuning a text-to-image diffusion model to an image-to-image model by substituting its text encoder with an image encoder. Subsequently, we fine-tune the model with explicit multi-view attention mechanism (EMA) which combines noisy multi-view images with the noise-free reference image as an explicit condition. CLIP embedding is sent to diffusion model throughout the whole process while reference images are discarded once after fine-tuning.
Extensive experimental results demonstrate that with a single image CLIP embedding, Isotropic3D is capable of generating multi-view mutually consistent images and a 3D model with more well-proportioned geometry, colored texture, and less distortion compared with existing image-to-3D methods while still preserving the similarity to the reference image as much as possible.



\bibliographystyle{splncs04}
\bibliography{main}


\clearpage

\appendix
\section*{\Large Appendix}



\section{Camera Model}
\label{sec:cameramodel}

We define $\theta$, $\varphi$ and $d$ to represent elevation angle, azimuth angle and camera distance respectively. 
We systematically sample camera viewpoints within the ranges $\theta \in [-10^\circ, 40^\circ]$, $\varphi \in [0^\circ, 360^\circ]$, and fix $d$ at 1.5. In practice, we convert these angles into radians. The positions of the camera for reference viewpoints and target views are denoted as $(\theta_r, \varphi_r, d_r)$ and $(\theta_n, \varphi_n, d_n)$, respectively, with $d = d_r = d_n = 1.5$ and $n \in \{1, 2, ..., N\}$ representing $N$ distinct target views. The relative transformation between the camera positions is captured by the expression ($(\theta_n - \theta_r, \varphi_n - \varphi_r, d)$). For the single-view diffusion branch, the target view can be any viewpoint. In contrast, during the multi-view diffusion procedure, the target views are confined to four distinct orthogonal viewpoints, each characterized by $\theta = 30^\circ$ and varying $\varphi \in [0^\circ, 360^\circ]$.

\section{Training Details}
\label{sec:train_detail}

{\bf View-Conditioned Multi-view Diffusion.} We take MVDream 
\cite{shi2023mvdream} as the base model for the first stage of fine-tuning.
In our optimization process, the AdamW optimizer is employed with a constant learning rate of $10^{-4}$ and a warm-up phase of 10,000 steps, complemented by the parameters $\beta_1 = 0.9$, $\beta_2 = 0.999$, and a weight decay of $10^{-2}$. We adopt gradient accumulation over 4 steps,  resulting in a total batch size of 768. This configuration allowed us to fine-tune the entirety of the UNet model across 50,000 iterations on eight A800 GPUs, spanning a total duration of approximately 7 days.
For the second stage, we continue utilizing the AdamW optimizer; however, we transition to a linear learning rate schedule that peaks at $5 \times 10^{-5}$. No gradient accumulation is used. We fine-tune the full UNet with a total batch size of 128 for 5k steps, which takes about 5 hours. Other settings are consistent with the first stage.

{\bf Isotropic3D. } Our 3D contents are optimized with 10,000 steps on a single A800 GPU. We adopt the AdamW optimizer at a learning rate of 0.01. 
For regularizing geometry, we choose point lighting and soft shading. We empirically set the guidance scale to 10 which is the same as during multi-view diffusion training. 
For SDS, we adjust the maximum and minimum time steps from 0.98 to 0.5 and 0.02 respectively in the 8,000 steps.
The coefficient of SDS loss is set as $\lambda_e = 1$ and the weight of orient loss can be written as
\begin{equation}
\lambda_o =\left\{
\begin{aligned}
& 0.2x, \quad x \leq 5,000, \\
& 1000, \quad 5000 < x \leq 10,000, \\
\end{aligned}
\right.
\end{equation}
where $x$ is the global step. Other settings not explicitly mentioned adhere to the specifications outlined in MVDream.

\section{Discuss Removing the Reference Image During Inference}

During training, the rationale behind concatenating the noise-free reference image and the noisy target images is to ensure consistency between the noisy targets and the reference image at the feature level. However, we aim to prevent the reference image from influencing the generation of multi-view images. Therefore, although the noise-free reference image is inputted into the network during training, we deliberately exclude it when calculating the loss with ground truth. It means that the reference image only gives explicit feature-level information to other target views in the self-attention layer without affecting the optimization of the model. Consequently, during the inference process, leveraging a single image CLIP embedding, our model is adept at generating multi-view images that are mutually consistent while still retaining a substantial degree of similarity to the reference image.

\section{Limitations and Discussion}
Although Isotropic3D has better geometry and texture than existing image-to-3D models and exhibits isotropy, the resolution of the rendered 3D contents is not high. We analyze that this may be because our training data only has a resolution of 256. After our extensive experimental results, our model does not perform well on faces, which may require further fine-tuning in downstream tasks.

\section{Addtional Results}
\textbf{Addtional results for qualitative comparisons}. In \cref{fig:more3d_comp}, we show more additional comparison results between Isotropic3D and other baselines. The video on the project website shows a more intuitive effect. 

\textbf{More results.} In order to verify the stability of the 3D generation, \cref{fig:more3d1} shows that two groups of results are from two runs with different seeds. For more detailed and vivid results, we show more examples of 3D contents generated by Isotropic3D in \cref{fig:more3d2} to \cref{fig:more3d8}. Please visit our website at \url{https://isotropic3d.github.io/}.

\begin{figure*}
  \centering
  \includegraphics[width=1.\linewidth]{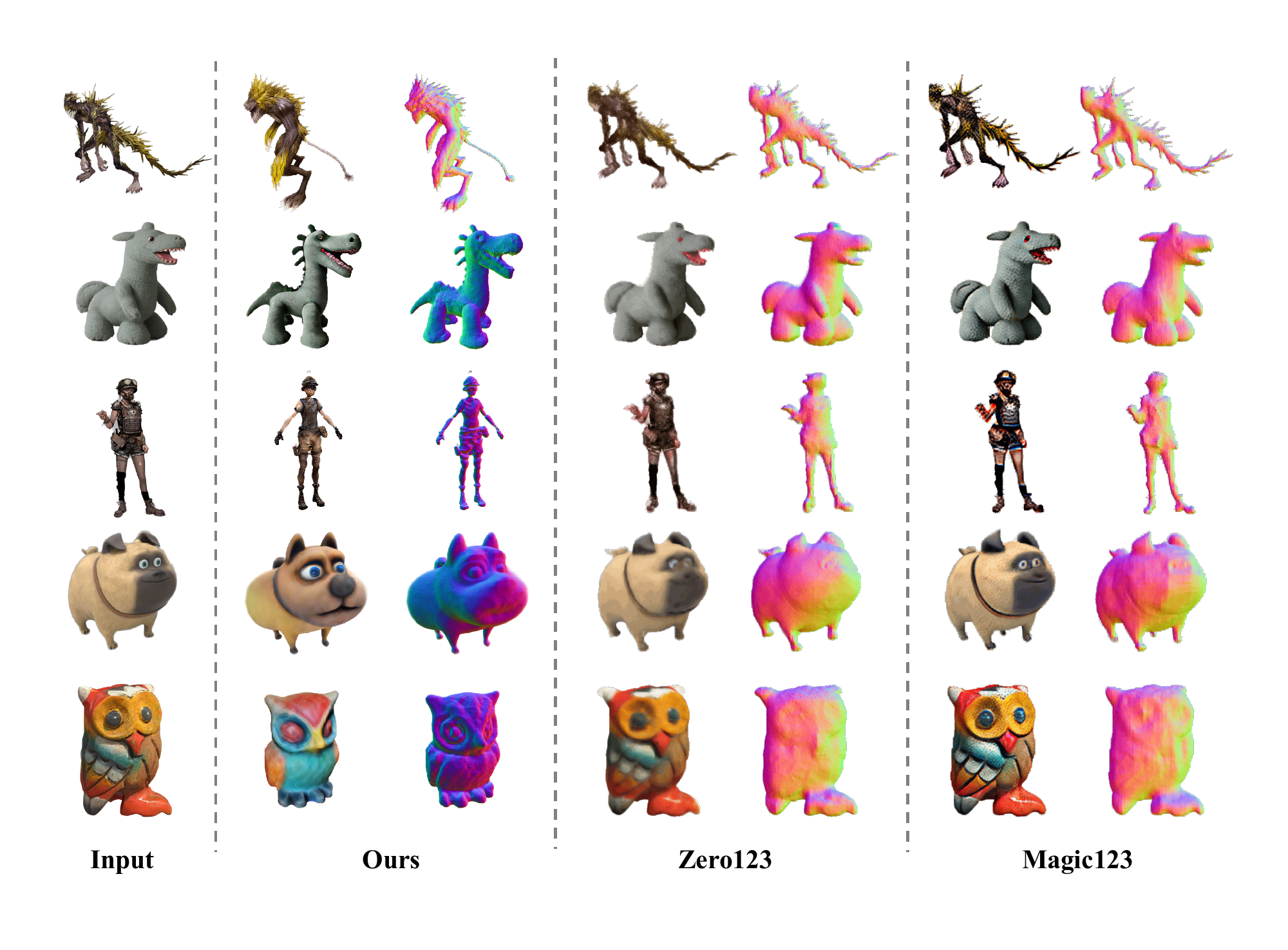}
  \caption{Addtional Results for qualitative comparisons of Isotropic3D with baseline models. A video of this result is available on our project website.}
  \label{fig:more3d_comp}
\end{figure*}

\begin{figure*}
  \centering
  \includegraphics[width=1.\linewidth]{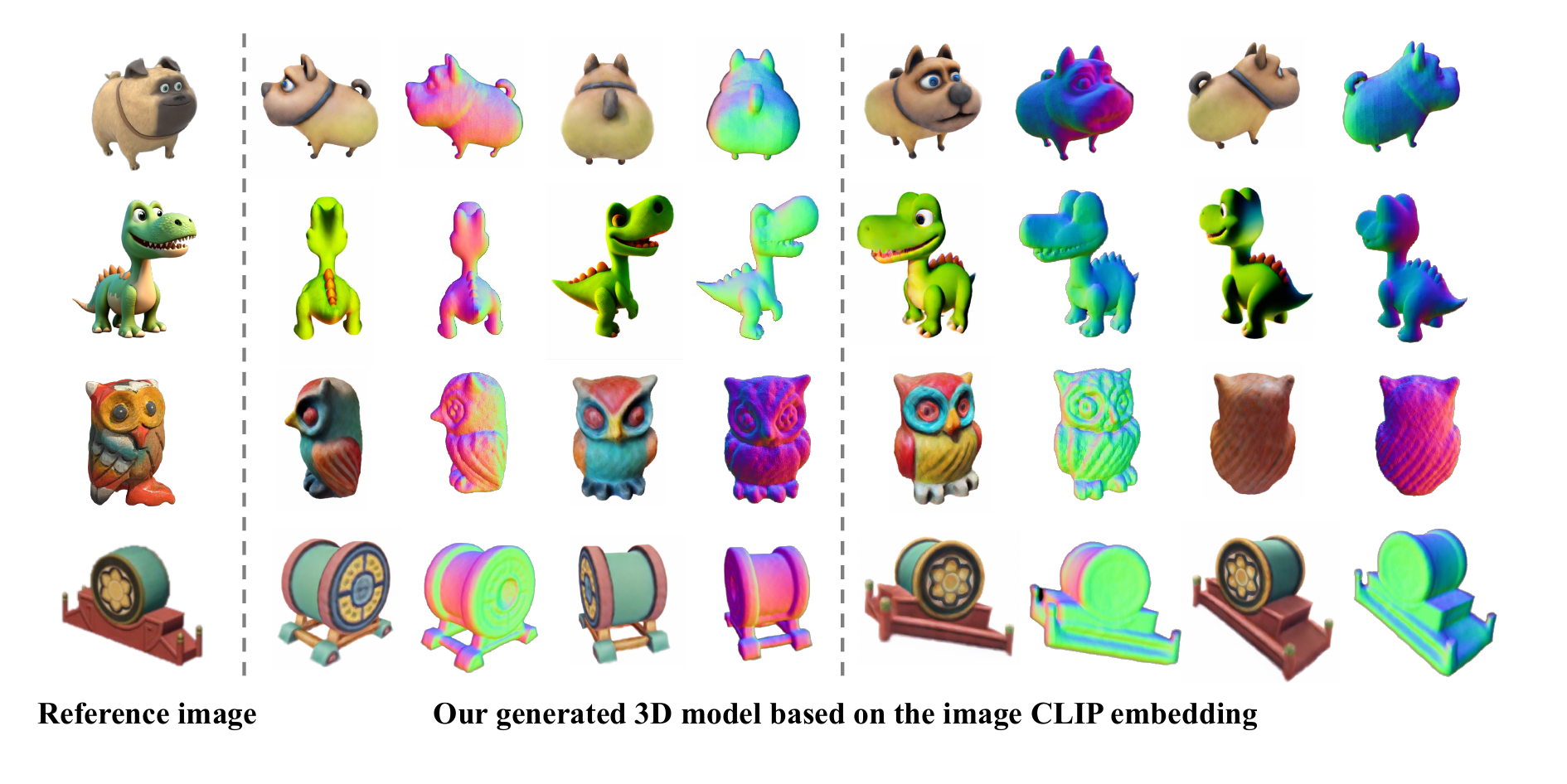}
  \caption{Addtional Results generated by {\bf Isotrpic3D}. Two groups of results are from two runs.}
  \label{fig:more3d1}
\end{figure*}

\begin{figure*}
  \centering
  \vspace{-0.3 in}
  \includegraphics[width=1.\linewidth]{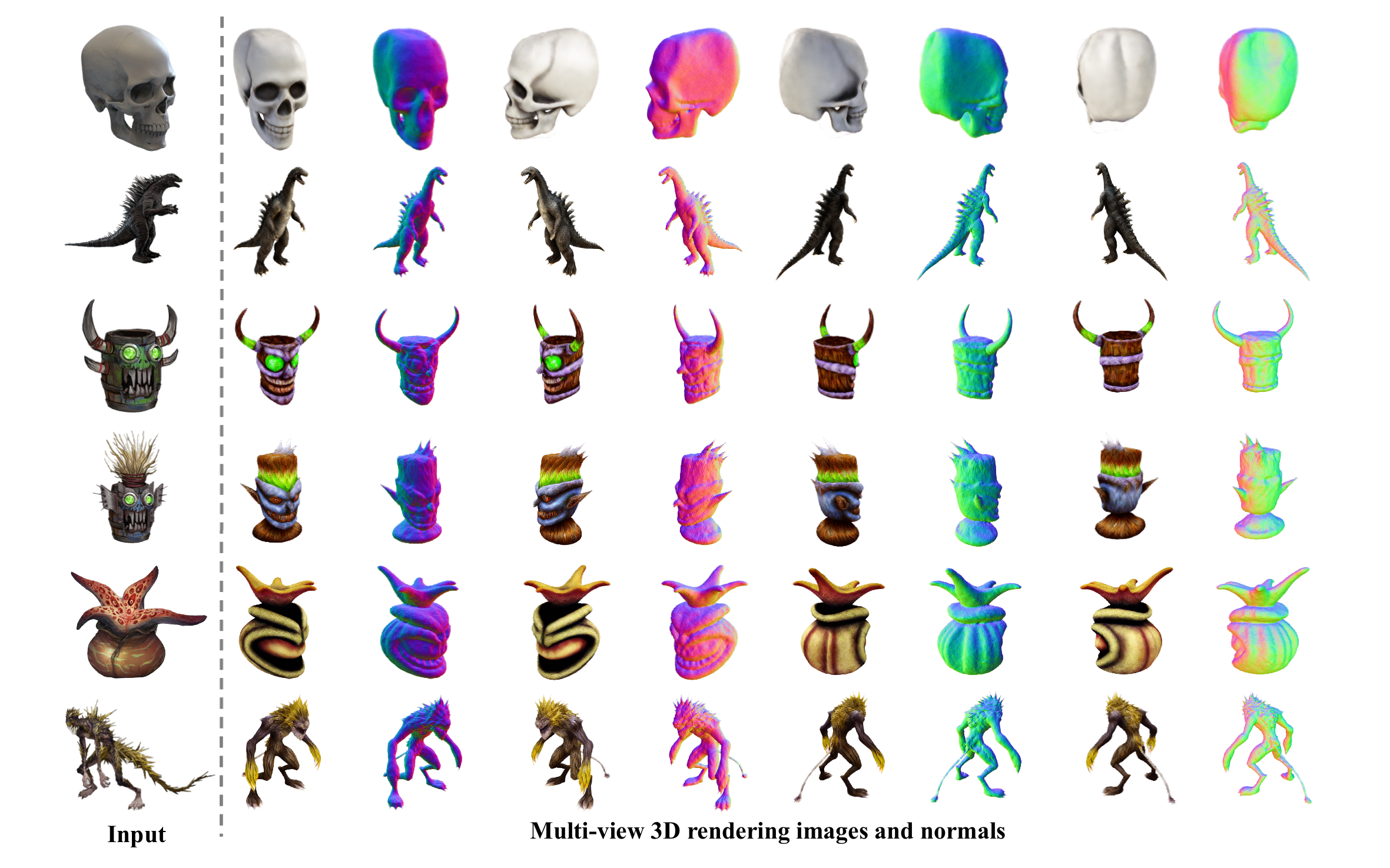}
  \caption{Example 3D contents generated by {\bf Isotrpic3D}. }
  \label{fig:more3d2}
\end{figure*}
\begin{figure*}
  \centering
  \vspace{-0.15 in}
  \includegraphics[width=1.\linewidth]{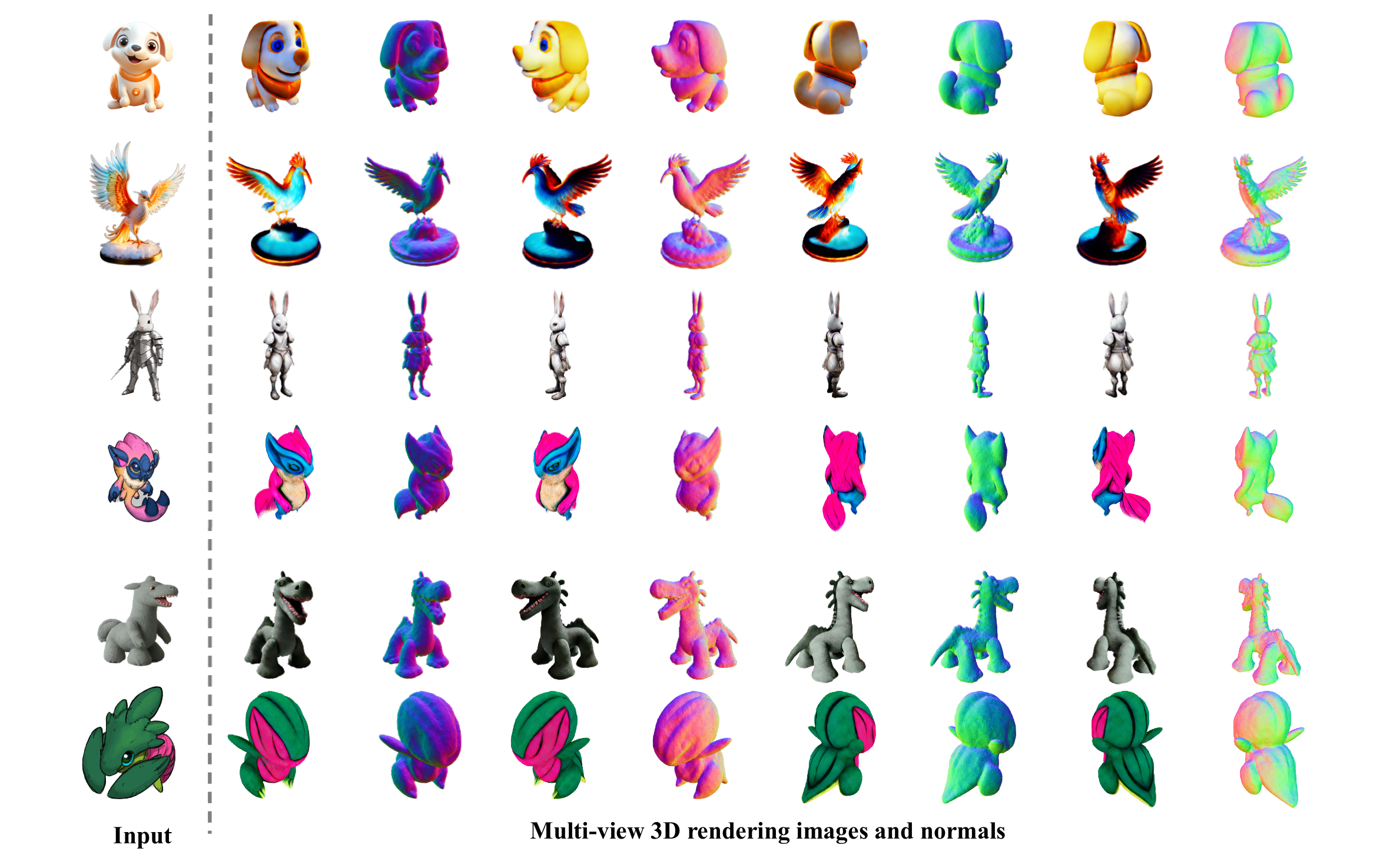}
  \caption{Example 3D contents generated by {\bf Isotrpic3D}.}
  \label{fig:more3d3}
\end{figure*}

\begin{figure*}
  \centering
  \vspace{-0.3 in}
  \includegraphics[width=1.\linewidth]{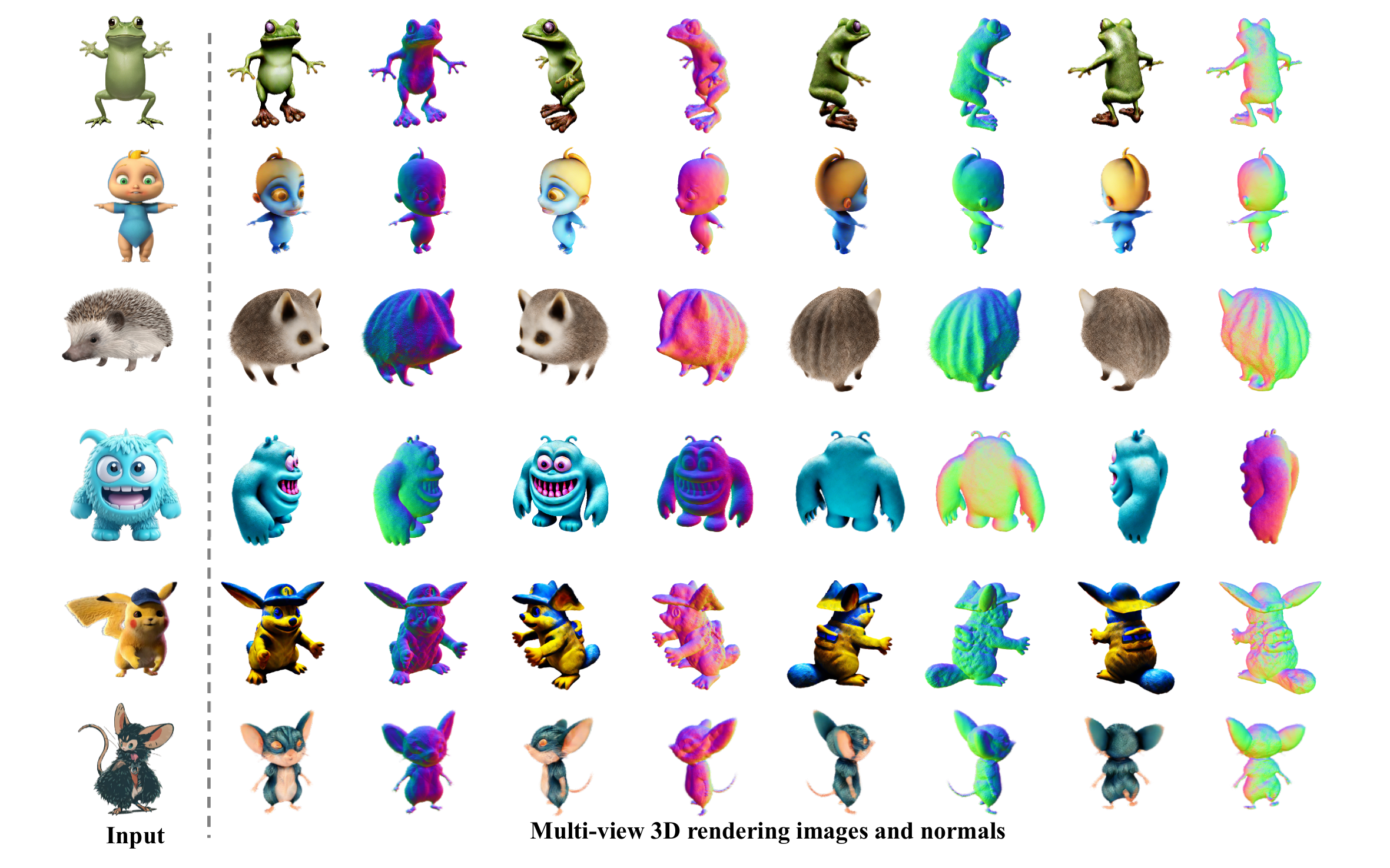}
  \caption{Example 3D contents generated by {\bf Isotrpic3D}.}
  \label{fig:more3d4}
\end{figure*}
\begin{figure*}
  \centering
  \vspace{-0.15 in}
  \includegraphics[width=1.\linewidth]{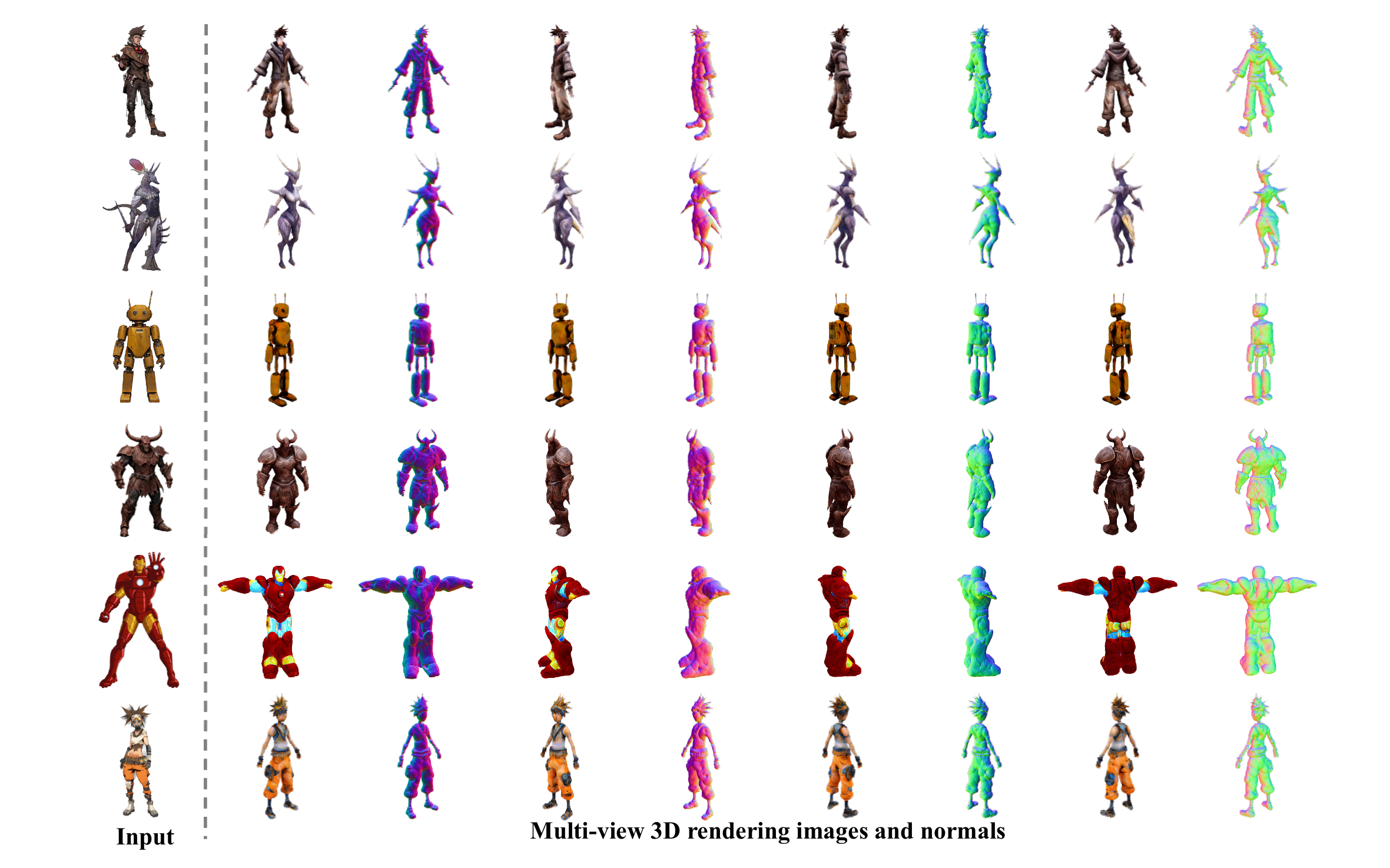}
  \caption{Example 3D contents generated by {\bf Isotrpic3D}.}
  \label{fig:more3d5}
\end{figure*}

\begin{figure*}
  \centering
  \vspace{-0.3 in}
  \includegraphics[width=1.\linewidth]{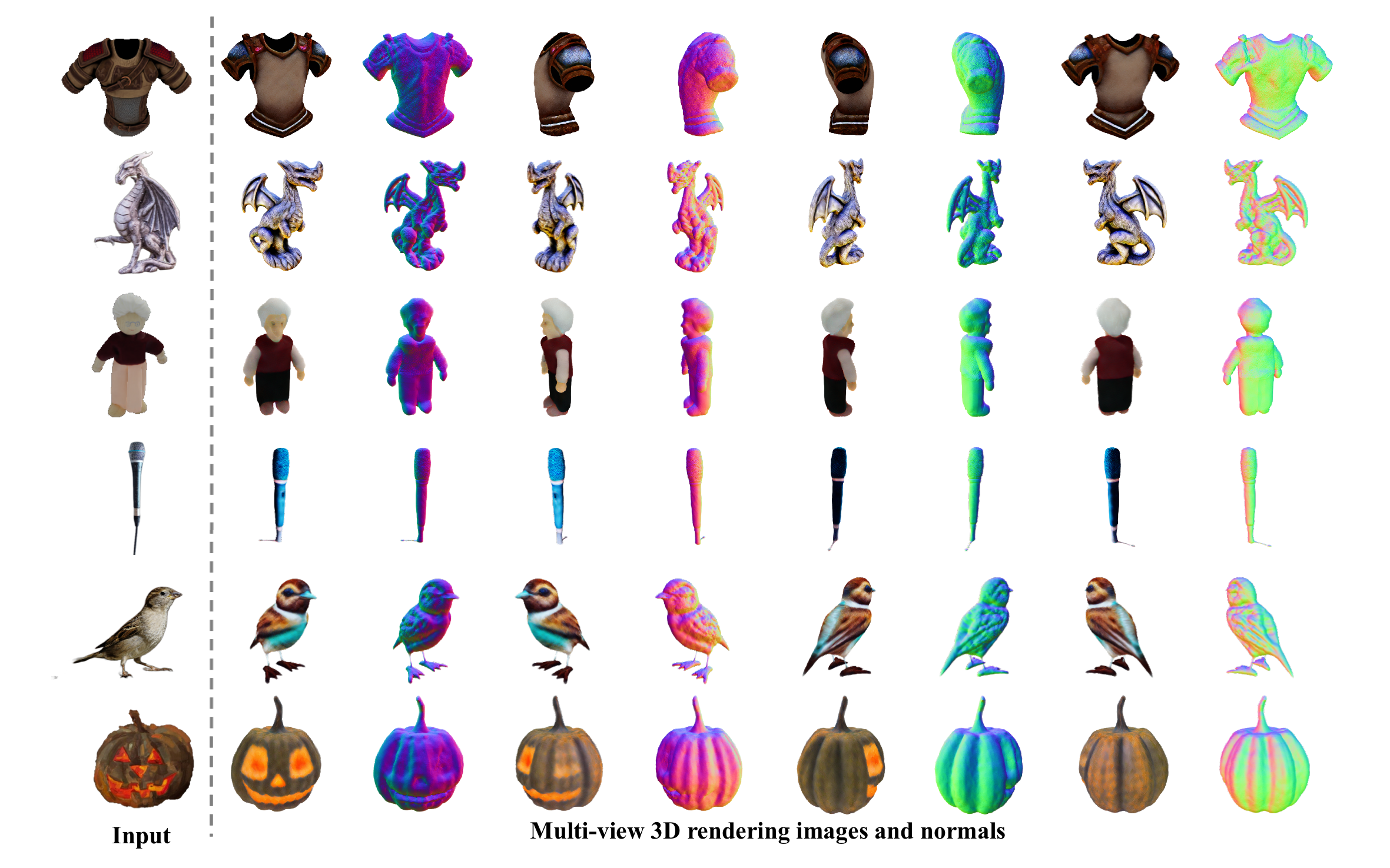}
  \caption{Example 3D contents generated by {\bf Isotrpic3D}.}
  \label{fig:more3d6}
\end{figure*}
\begin{figure*}
  \centering
  \vspace{-0.15 in}
  \includegraphics[width=1.\linewidth]{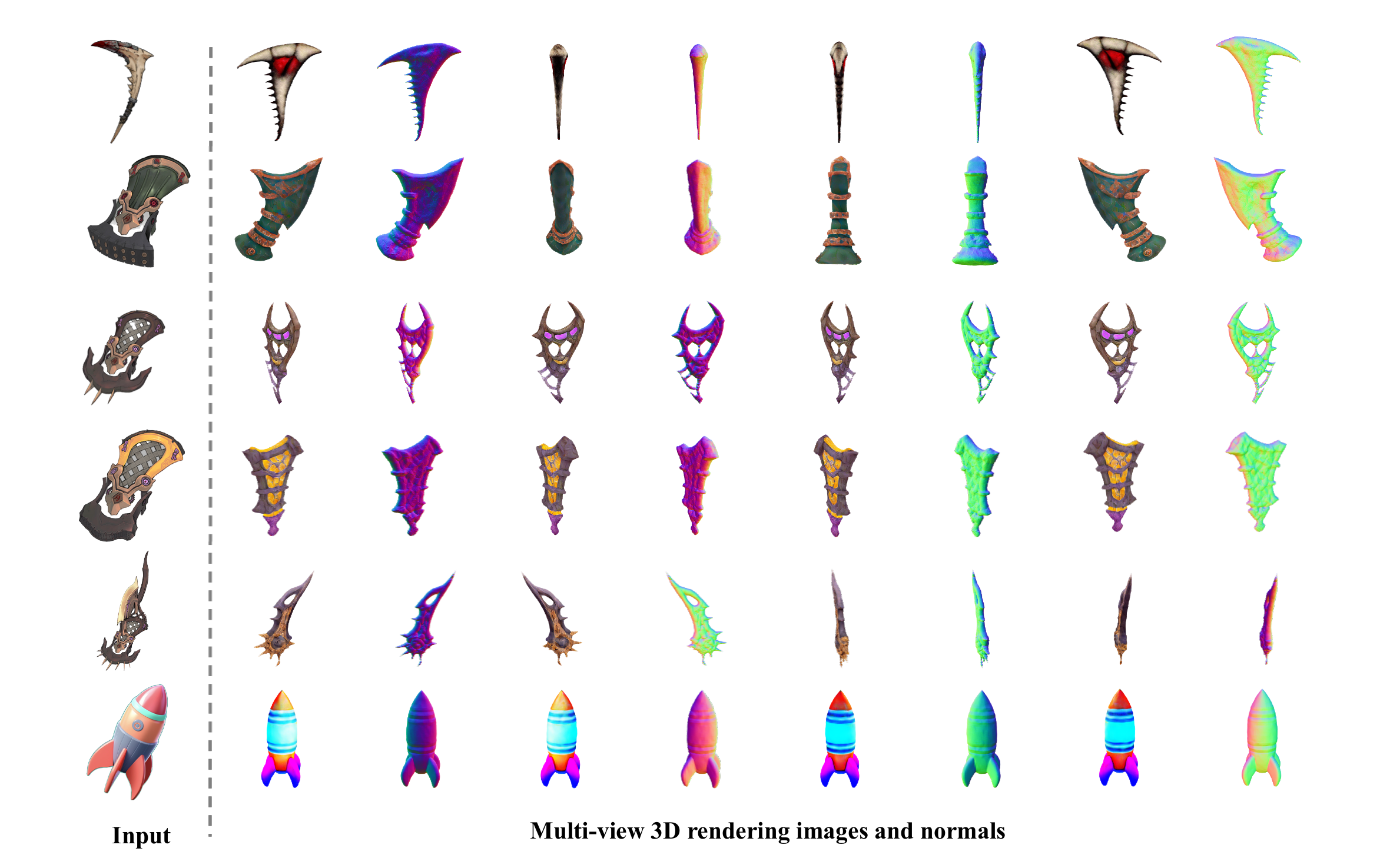}
  \caption{Example 3D contents generated by {\bf Isotrpic3D}.}
  \label{fig:more3d7}
\end{figure*}
\begin{figure*}
  \centering
  \vspace{-0.3 in}
  \includegraphics[width=1.\linewidth]{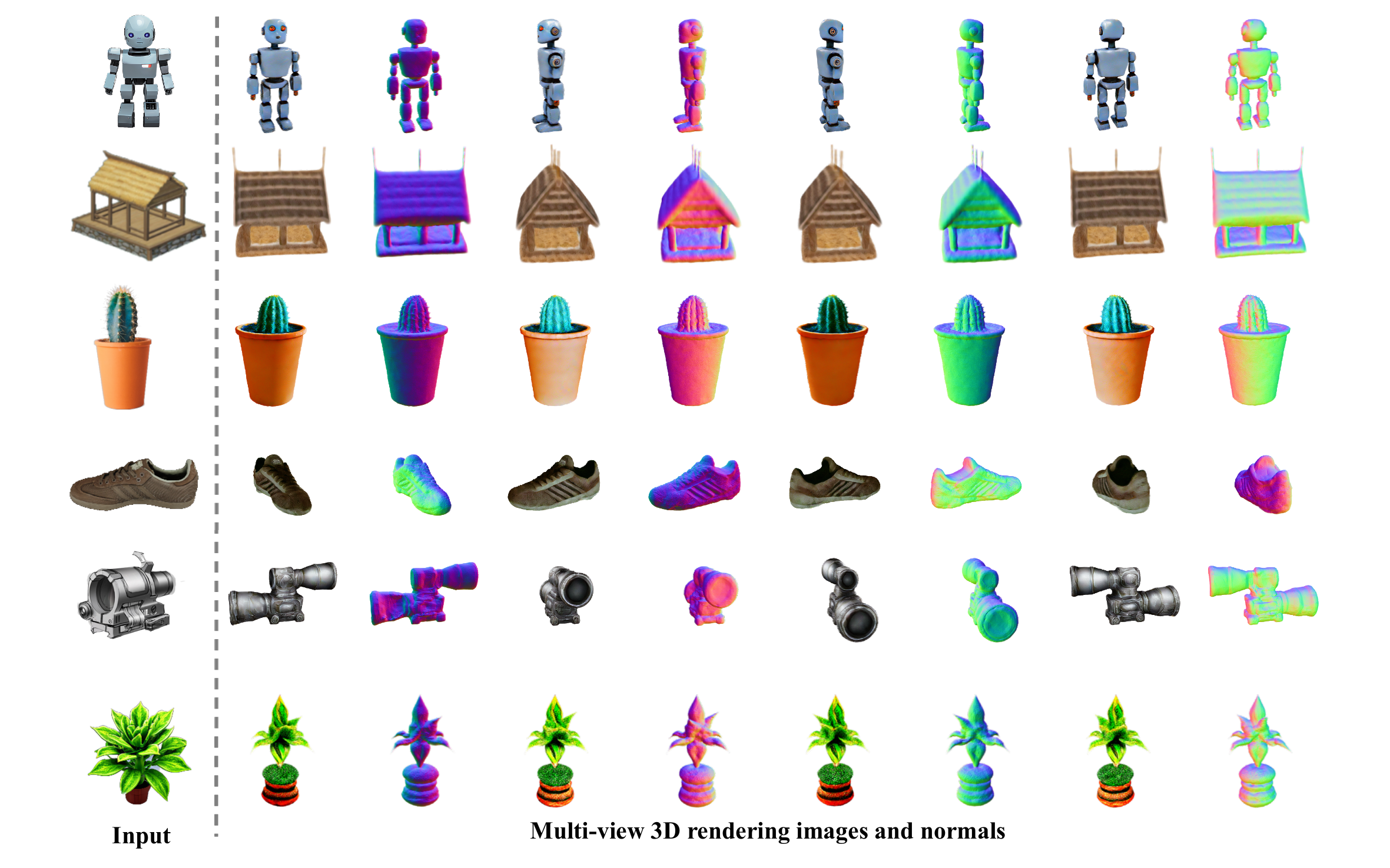}
  \caption{Example 3D contents generated by {\bf Isotrpic3D}.}
  \label{fig:more3d8}
\end{figure*}

\end{document}